\documentclass[10pt,letterpaper]{article}
\usepackage[top=0.85in,left=2.75in,footskip=0.75in]{geometry}

\usepackage{amsmath,amssymb}

\usepackage{changepage}

\usepackage[utf8x]{inputenc}

\usepackage{textcomp,marvosym}

\usepackage{cite}

\usepackage{nameref,hyperref}

\usepackage{multirow}

\usepackage{graphicx}
\graphicspath{figures}

\usepackage{subcaption}

\usepackage{microtype}
\DisableLigatures[f]{encoding = *, family = * }

\usepackage[table]{xcolor}

\usepackage{array}

\newcolumntype{+}{!{\vrule width 2pt}}

\newlength\savedwidth

\newcommand\thickhline{\noalign{\global\savedwidth\arrayrulewidth\global\arrayrulewidth 2pt}%
\hline
\noalign{\global\arrayrulewidth\savedwidth}}

\raggedright
\usepackage[aboveskip=1pt,labelfont=bf,labelsep=period,justification=raggedright,singlelinecheck=off]{caption}

\makeatletter
\renewcommand{\@biblabel}[1]{\quad#1.}
\makeatother

\usepackage{lastpage,fancyhdr,graphicx}
\usepackage{epstopdf}
\fancyheadoffset[L]{2.25in}
\fancyfootoffset[L]{2.25in}
\begin{document}
\vspace*{0.2in}

\begin{flushleft}
{\Large
\textbf\newline{Semantic and structural image segmentation for prosthetic vision} % Please use "sentence case" for title and headings (capitalize only the first word in a title (or heading), the first word in a subtitle (or subheading), and any proper nouns).
}
\newline
\\
Melani Sanchez-Garcia\textsuperscript{1*},
Ruben Martinez-Cantin\textsuperscript{1,2},
Jose J. Guerrero\textsuperscript{1}
\\
\bigskip
\textbf{1} Instituto de Investigaci\'{o}n en Ingenier\'{i}a de Arag\'{o}n (I3A). Universidad de Zaragoza, Spain
\\
\textbf{2} Centro Universitario de la Defensa, Zaragoza, Spain
\\
\bigskip

% Insert additional author notes using the symbols described below. Insert symbol callouts after author names as necessary.
% 
% Remove or comment out the author notes below if they aren't used.
%
% Primary Equal Contribution Note
%\Yinyang These authors contributed equally to this work.

% Additional Equal Contribution Note
% Also use this double-dagger symbol for special authorship notes, such as senior authorship.
%\ddag These authors also contributed equally to this work.

%\textcurrency Current Address: Dept/Program/Center, Institution Name, City, State, Country % change symbol to "\textcurrency a" if more than one current address note
% \textcurrency b Insert second current address 
% \textcurrency c Insert third current address


% Group/Consortium Author Note
%\textpilcrow Membership list can be found in the Acknowledgments section.

% Use the asterisk to denote corresponding authorship and provide email address in note below.
* mesangar@unizar.es

\end{flushleft}
\section*{Abstract}

Prosthetic vision is being applied to partially recover the retinal stimulation of visually impaired people. However, the phosphenic images produced by the implants have very limited information bandwidth due to the poor resolution and lack of color or contrast. The ability of object recognition and scene understanding in real environments is severely restricted for prosthetic users. Computer vision can play a key role to overcome the limitations and to optimize the visual information in the simulated prosthetic vision, improving the amount of information that is presented. We present a new approach to build a schematic representation of indoor environments for phosphene images. The proposed method combines a variety of convolutional neural networks for extracting and conveying relevant information about the scene such as structural informative edges of the environment and silhouettes of segmented objects. Experiments were conducted with normal sighted subjects with a Simulated Prosthetic Vision system. The results show good accuracy for object recognition and room identification tasks for indoor scenes using the proposed approach, compared to other image processing methods.

% Please keep the Author Summary between 150 and 200 words
% Use first person. PLOS ONE authors please skip this step. 
% Author Summary not valid for PLOS ONE submissions.   
%\section*{Author summary}
%Lorem ipsum dolor sit amet, consectetur adipiscing elit. Curabitur eget porta erat. Morbi consectetur est vel gravida pretium. Suspendisse ut dui eu ante cursus gravida non sed sem. Nullam sapien tellus, commodo id velit id, eleifend volutpat quam. Phasellus mauris velit, dapibus finibus elementum vel, pulvinar non tellus. Nunc pellentesque pretium diam, quis maximus dolor faucibus id. Nunc convallis sodales ante, ut ullamcorper est egestas vitae. Nam sit amet enim ultrices, ultrices elit pulvinar, volutpat risus.

%\linenumbers

% Use "Eq" instead of "Equation" for equation citations.
\section*{Introduction}

\begin{figure}[!h]
%\begin{adjustwidth}{-2.25in}{0in} 
\centering
\includegraphics[width=1\textwidth]{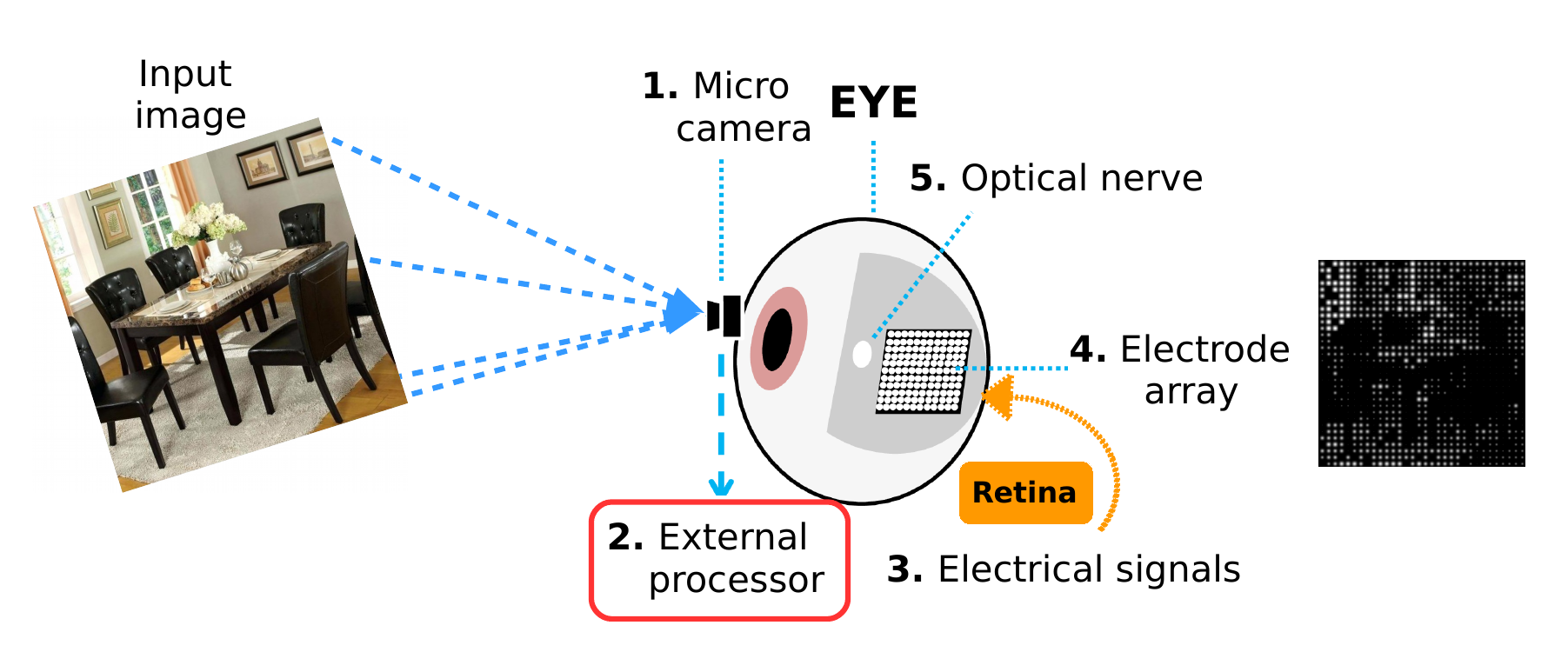}
\caption{{\bf Configuration of a retinal prosthesis.} 
The external and internal components include a micro camera, a transmitter, a external processing unit and a implanted electrode array. First, the external camera acquires an image. Then, the external processor converts the image to a suitable pattern of electrical stimulation of the retina through an electrode array.}
\label{Fig1}
%\end{adjustwidth}
\end{figure}

Retinal degenerative diseases such as retinitis pigmentosa and age-related macular degeneration cause loss of vision due to the gradual degeneration of the sensory cells in the retina. Visual prosthesis are currently the most promising technology to improve vision in patients with such degenerative diseases. These devices elicit visual perception by electrically stimulating retina cells. As a result, implanted patients are able to see patterns of spots of light called \emph{phosphenes} that the brain interprets as a visual information. However, retinal implants are limited to hundreds of electrical receptors, which produce a very limited visual elicitation. From the actual technologies for retinal implants \cite{kien2012review}, one of the most active line of research is based on implants with a micro camera that captures external stimuli and a processor that converts the visual information in microstimulations in the implant, as can be seen in Fig~\ref{Fig1}. Following the computer image paradigm, we can say that the visual information evoked by the implants has very low spatial resolution and very limited dynamic range (only few levels of stimulus intensity are perceived as different) \cite{eiber2013attaining,chen2007quantitative,dagnelie2006visual}. Intuitively, from an information theory perspective, the process from the external sensor input to the implant stimuli is analogous to taking a high definition image and convert it to a low resolution, grayscale image with just a few grey levels. Thus, a large amount of visual information is lost. Prosthetic vision allows users to recognize objects with simple shapes, to see people’s silhouettes in bright light or detect motion \cite{humayun2003visual}, but high level tasks require more precise visual cues and a deeper interpretation of the information.

Recent developments in implants might result in an improved resolution and performance of the visual elicitation \cite{choi2017human}, but quality would still be several orders of magnitude lower than a current micro camera. Alternatively, the visual information gathered by the external camera could be processed prior to being transferred to the retinal electrodes. Image processing can be used to extract and highlight relevant information from the external camera. This information can be presented with visual cues that help to understand the perceived scene by the implanted subject. Several studies have already been conducted testing specific cues for object recognition \cite{hayes2003visually,mace2015simulated,qiu2018motion,wang2014moving,guo2018optimization,li2018image,van2010simulating}, reading \cite{bourkiza2013visual,sommerhalder2003simulation,dagnelie2006paragraph,vurro2014simulation}, facial recognition \cite{mckone2018caricaturing, wang2014face} or navigation \cite{vergnieux2017simplification,dagnelie2007real,vergnieux2014wayfinding,perez2017depth,parikh2013performance} in the context of prosthetic vision.

One of the most basic image processing tasks from the cognitive, but also from the computational level, is the segmentation of the image in different regions \cite{zhu2016beyond}. From a statistical point of view, this corresponds to the problem of \emph{clustering}. Rooted on the Aristotelian laws of association, early research in perception from \emph{Gestalt} psychologists found the importance of the \emph{principles of grouping} \cite{wertheimer1938laws}. These principles state that our brain tends to group image elements based on proximity, color, shape or other similarities. Although some of the Gestalt ideas are controverted, the principles of grouping have been supported by posterior empirical research \cite{bruce2003visual,hoffman1997salience}. From a computational perspective, image segmentation dates back to the seminal work of Minsky and Pappert \cite{minsky1967mit} followed by several works in the 60s and 70s \cite{brice1970ai,haralick1985image}. However, segmentation is still an open problem with an active research community thanks to recent outstanding results in segmentation with semantic meaning \cite{garcia2017review}. In semantic segmentation, the objective is to group the image regions based on labels with semantic meaning. This transforms the clustering problem into a classification problem. In fact, recent research has gone one step further and provide \emph{instance-aware} semantic segmentation \cite{dai2016instance} which is able to group pixels of single objects, separated from other objects even if they belong to the same semantic class. 

Current state of the art methods for image segmentation are mostly based on deep neural networks \cite{garcia2017review}. Most recent developments of semantic and instance-aware segmentation are based on Fully Convolution Networks (FCN) \cite{dai2016instance,fernandez2018panoroom,he2017mask,garcia2017review, mallya2015learning}. They are an architecture based on convolutional layers with added upsampling layers with skip connections to allow for detailed pixel prediction on arbitrary-sized inputs \cite{long2015fully}. Others approaches, like the U-net architecture, are able to provide accurate pixel prediction \cite{ronneberger2015u}.

In this work, we use semantic segmentation to enhance the visual stimuli for accurate indoor scene understanding using visual prosthesis. Basic scene understanding is the stepping stone for many everyday tasks for visually impaired people and also facilitates subsequent tasks of finer perception. Concretely, we use two different types of semantic segmentation based on FCNs to highlight the information available in the image and to present the most useful information to the user. We use instance-aware semantic segmentation to group the pixels of relevant objects in the scene. This has a double purpose: on one hand, we are able to reduce visual clutter, which becomes indistinguishable noise in low resolution implant array; on the other hand, the grouping highlights the silhouette of the object, making it more distinguishable. One of the main problems of using object silhouettes for recognition is the lack of sense of scale or perspective. Thus, we rely on a second semantic segmentation network to extract structural informative edges of the scenes, such as wall and ceiling intersections. Those edges provide an intuitive representation of the 3D structure of the room \cite{Sanchez-Gracia19visapp}. We evaluate and compare the proposed semantic and structural image segmentation with baseline methods through a Simulated Prosthetic Vision (SPV) experiment, which is a standard procedure for non-invasive evaluation using normal vision subjects \cite{hayes2003visually,mace2015simulated,qiu2018motion,wang2014moving,guo2018optimization,li2018image,van2010simulating,bourkiza2013visual,sommerhalder2003simulation,dagnelie2006paragraph,vurro2014simulation,mckone2018caricaturing,wang2014face,vergnieux2017simplification,dagnelie2007real,vergnieux2014wayfinding,perez2017depth,parikh2013performance}. The experiments included two tasks: object recognition and room identification.

\section*{Methods}

\subsection*{Subjects}

Eighteen subjects with normal vision volunteered for the formal experiment. The subjects (four females and fourteen males) were between 20 and 57 years old. 

\subsubsection*{Ethics Statement}

The research process was conducted according to the ethical recommendations of the Declaration of Helsinki and was approved by the Aragon Autonomous Community Research Ethics Committee (CEICA) that evaluates human research projects, human biological samples or personal data. The research protocol used for this study is non-invasive, purely observational, with absolutely no-risk for any participant. There is no personal data collection or treatment and all subjects were volunteers. Subjects gave their informed written consent after explanation of the purpose of the study and possible consequences. The consent allowed the abandonment of the study at any time. All data were analyzed anonymously.

\subsection*{Stimuli}
\label{sec:ImageGeneration}

%The image processing stage in visual prosthesis is the first step to set the resolution of input image corresponding to the number of stimulating electrodes. The limited number of electrodes in the current implants can lead to huge loss of information when presenting the daily scenes. This severely restricts the ability of prosthetic user to recognize objects and scenes. Segmenting the image into semantically significant parts can optimize the visual information presentation in simulated vision. Therefore, automatically detecting the main elements and precisely separating the objects from the scenes are needed first. Due to the recent advances in deep learning for semantic segmentation 

\begin{figure}[!h]
% \begin{adjustwidth}{-2.25in}{0in} 
\centering
\includegraphics[width=\textwidth]{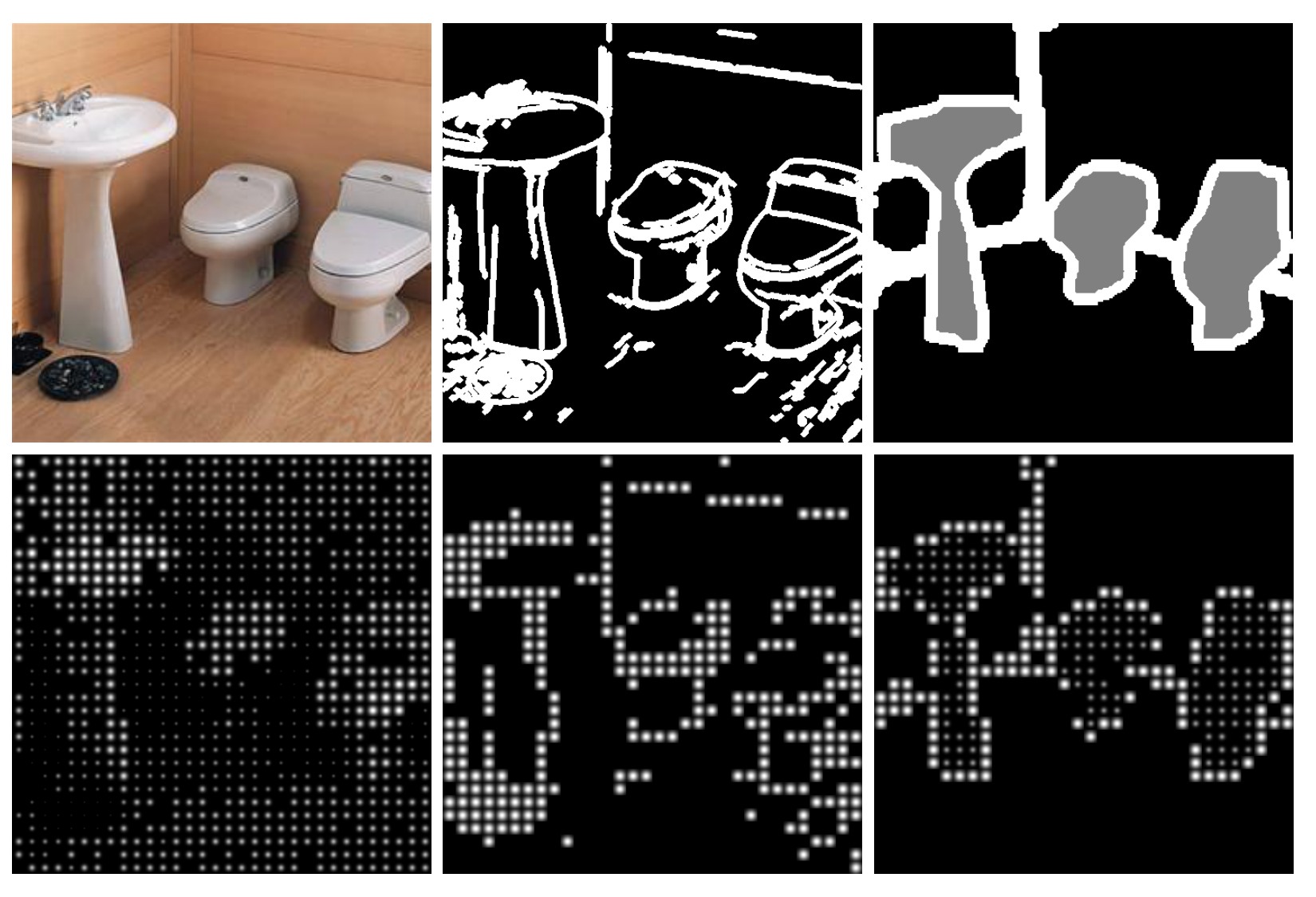}
\caption{{\bf Stimuli generation.} Top row: Example of a bathroom scene with the three processing methods used in this work (a) Direct image, (b) Edge image and (c) SIE-OMS image. Bottom row: the three processing methods in the SPV.} 
\label{Fig2}
% \end{adjustwidth}
\end{figure}

We use Simulated Prosthetic Vision which allows to represent the descriptions of phosphene perception reported by visual prosthesis patients. For our phosphenic image generation after the stimuli, the array of phosphenes was limited to 32 x 32 (1024 electrodes) and 8 different luminance levels according to the number of luminance levels attainable in human trials using retinal prostheses \cite{wang2014moving,chen2009simulating}. We also included a 10\% dropout of electrodes. The complete process of phosphene generation can be found in the Supplementary material (see \nameref{S1_File}). Examples of the resulting effect are shown in Fig~\ref{Fig2}. All the phosphenic images used in the experiments can be found in the dataset available online\footnote{Image dataset: \url{http://webdiis.unizar.es/~rmcantin/index.php/Research/SIEOMS}}. The following sections describe the three methods used to activate the phosphenes during the experiments. First, our proposal based on semantic segmentation (SIE-OMS). Then, two baseline methods: a) detecting the silhouettes and structure within the scene with a standard edge detector (Edge), and b) generating the stimulus directly from the input image luminance (Direct).

\subsubsection*{SIE-OMS}

\begin{figure}[!h]
\begin{adjustwidth}{-2.25in}{0in} 
\centering
\includegraphics[width=1.4\textwidth]{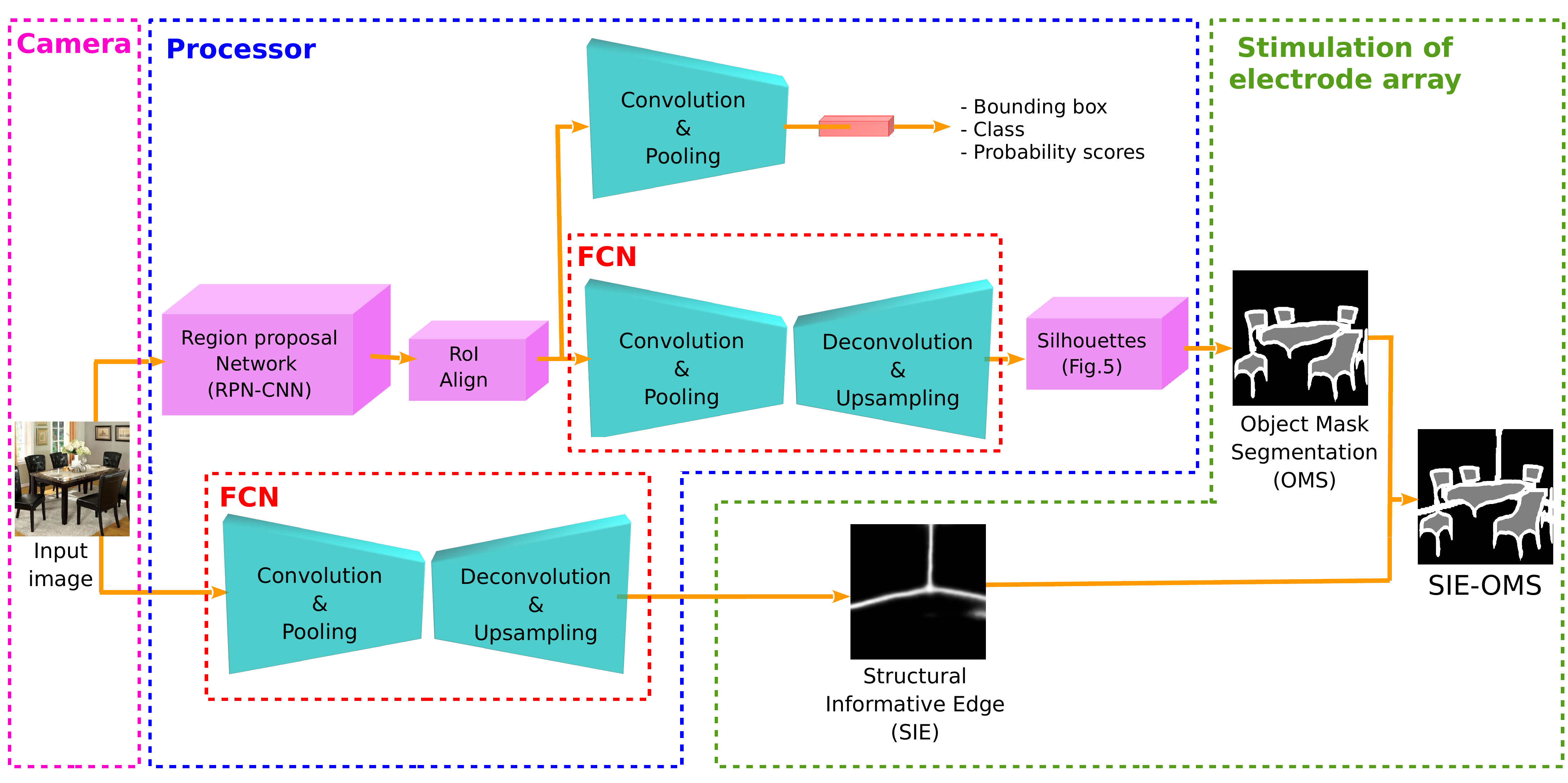}
\caption{{\bf Processing pipeline.} The stimulation of the electrode array is based on two information pathways to extract the regions of pixels that represents important objects (OMS) and structural edges (SIE). The regions are computed using two different types of FCN.}
\label{Fig3}
\end{adjustwidth}
\end{figure}

We propose to combine two FCNs to select and highlight informative elements in indoor scenes as an intelligent way of activating the phosphenes. Specifically, we extract structural informative edges (SIE) and object masks and silhouettes (OMS) to later combined both, SIE and OMS, to build our proposed schematic representation of the scene (SIE-OMS), as can be seen in Fig \ref{Fig3}.

\paragraph*{Structural informative edges (SIE)}

\begin{figure}[!h]
\centering
\includegraphics[width=1\textwidth]{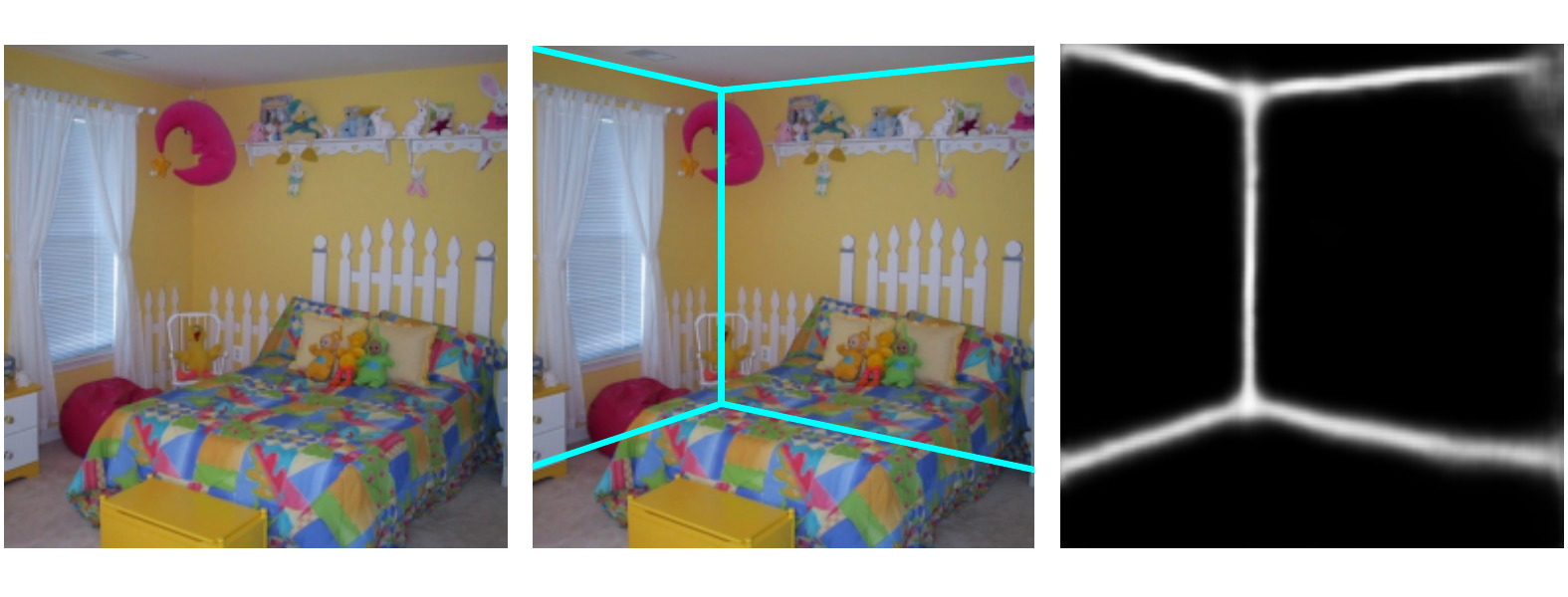}
\caption{{\bf Indoor scene layout from an indoor image.} 
Using \cite{fernandez2018panoroom} we detect the main structure of the room extracting the structural informative edges (SIE) (right) which are those formed by the intersection of walls, ceiling and floor of the room (middle).}
\label{Fig4}
\end{figure}

One of the main problems in the recognition of scene elements based on silhouettes is the lack of sense of scale or perspective. The scale and the structure of the scene can be achieved by detecting the structural informative edges (SIE), that is, those main edges formed by the intersection of the walls, floor and ceiling of the room. These edges can be seen in Fig~\ref{Fig4}. Our approach is based on the model by Fernandez-Labrador et al. \cite{fernandez2018panoroom} for indoor scenes. Similarly to the object masks network described below, this method is also based on a FCN for pixel classification. In this case, the network was trained to estimate probability maps representing the room structural edges, even in the presence of clutter and occlusions. We have used a model pretrained with the LSUN dataset \cite{yu2015lsun}.

\paragraph*{Object Masks and Silhouettes (OMS)}

\begin{figure}[!h]
\begin{adjustwidth}{-2.25in}{0in} 
\centering
\includegraphics[width=1.0\textwidth]{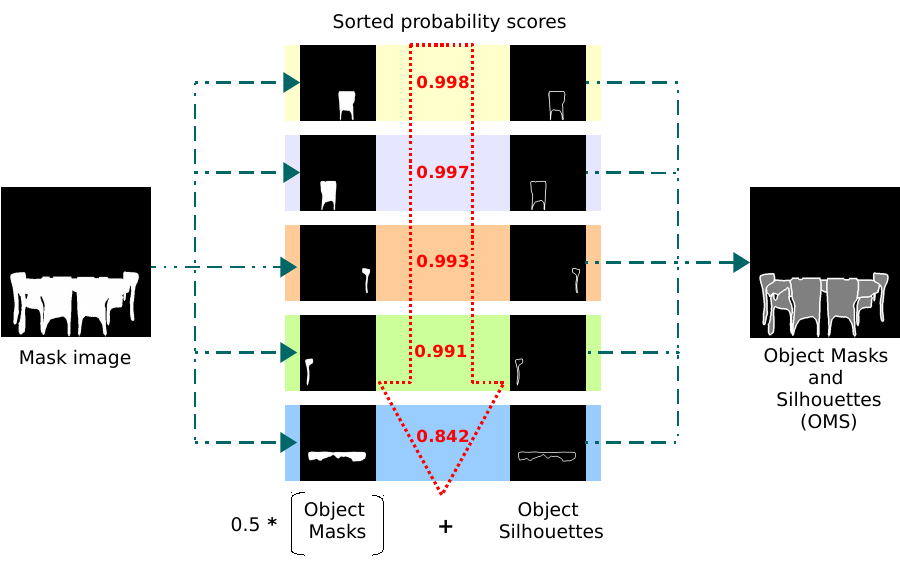}
\caption{{\bf Objects Masks and Silhouettes (OMS)}. Object masks were generated from \cite{he2017mask} and were sorted by probability scores to avoid occlusions between objects. The extracted information was combined in an image highlighting the silhouettes of the objects in white with the object masks in gray.}
\label{Fig5}
\end{adjustwidth}
\end{figure}

We perform instance segmentation of objects using the architecture of Mask R-CNN \cite{he2017mask} which is partially represented in Fig~\ref{Fig3}. The first part of the network, called a Region Proposal Network (RPN), proposes candidates about the regions that contain objects on the input image. The second module, called RoIAlign, runs on the regions of interest (ROIs) proposed by the RPN and aligns those regions to the feature maps extracted by the RPN. Then, the model splits in two branches. The box branch based on Faster R-CNN. It generates two outputs for each ROI: the class of the object in the ROI and a bounding box refinement of the object area using a regression model. The mask branch is a convolutional network that takes the positive regions selected by the ROI classifier and generates binary masks. Then, it uses up sampling to scale the predicted masks to the size of the ROI bounding box which gives the final masks, one per object. For our experiments, we have used a pre-trained model on the COCO dataset \cite{he2017mask,lin2014microsoft}. However, we have removed some of the object classes such as small objects (e.g.: scissors, banana, etc.) and non-indoor objects (e.g.: car, tree, etc.). As a final step, we highlighted the silhouettes of the objects to distinguish the masks of some overlapping objects, as can be seen in Fig~\ref{Fig5}. We also performed morphological operations to reduce the aliasing effect when translated to phosphene images.

\paragraph*{Dealing with occlusions}

Although this algorithm has achieved good results for object segmentation, there are more complicated cases, such as images with overlapping objects or scenes with occlusions, where the view of one object may be blocked by other objects. In that case, we could use a depth sensor, such as an RGB-D camera, or a stereo camera to estimate the depth. Alternatively, there are some works to estimate the depth purely, based on monocular information \cite{facil2017single}. As a proof of concept, we found that the probability score for the detection network was correlated to the level of occlusion of each object. That is, a high probability is most likely to appear in objects that are in the front. Thus, we stacked the instances from the least to the highest probability, leaving the objects with the highest score overlapping the objects with the least score, as can be seen in Fig~\ref{Fig5}. Finally, the SIE was always in the background.

\subsubsection*{Baseline methods}

We compared SIE-OMS with two baseline methods used in visual prosthesis: a) a direct method that converts the input image directly to the phosphene map by averaging the brightness on the region covered by each phosphene, and b) a standard edge detector to extract brightness contours (see Fig~\ref{Fig2}). The direct method has proved to be very effective in scenes where high contrast predominates \cite{barnes2012role}. Edge detectors have also been previously used for prothesis vision and phosphene images \cite{feng2013enhancing,snaith1998low,chang2012facial}. Since the contours of an image holds much information, edge extraction is a useful method of encoding and selecting the information contained in an image. The drawback here is that the understanding of a complete scene in low vision represented by edges may be more challenging because the amount of clutter. For example, Sanocki et al. investigated how complicated is an edge extractor method comparing object recognition with and without removal of background clutter with edge images \cite{sanocki1998edges}. The results showed that the increase in the number of edges greatly increase the complexity. For the edge detector, we used the Canny implementation from the \texttt{scikit image} Python package with the default parameters \cite{scikit-image}. In this case, we also added morphological operations (dilation) to reduce aliasing without adding clutter.

\subsection*{Experimental setup}

\begin{figure}[!h]
\centering
\includegraphics[width=0.8\textwidth]{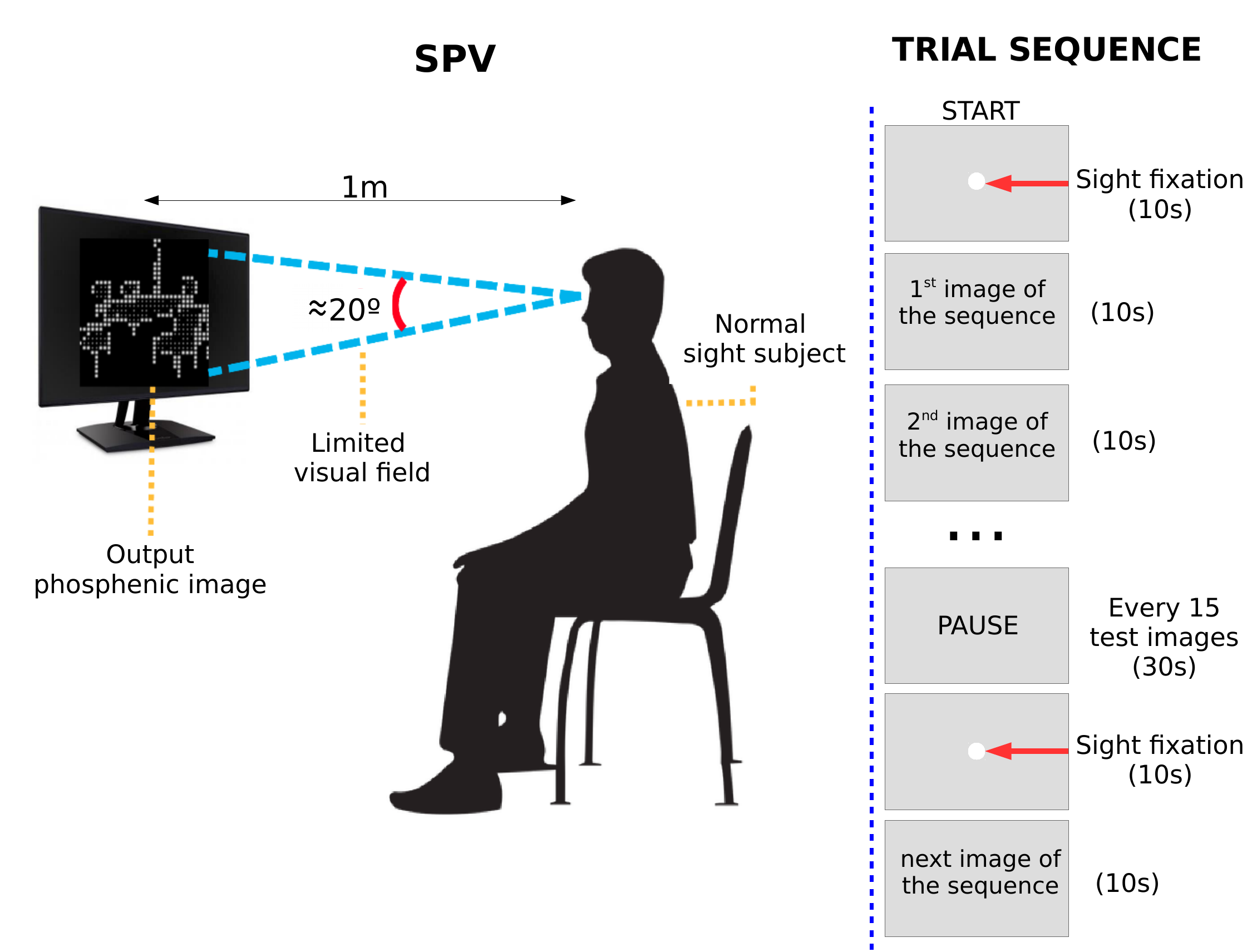}
\caption{{\bf SPV and trial setup.} 
SPV setup: Subjects were seated on a chair facing a computer screen at 1m distance. The visual field was 20 degrees that simulates the prosthesis device. Trial setup: Each gray rectangle represents the image shown on the computer monitor during the trial. Each image appeared for 10 seconds and switched for the next image automatically. Break time between image sequences was 30 seconds. The complete experiment took approximately 15 minutes.}
\label{Fig6}
\end{figure}

Most of the SPV configurations are usually based on a computer screen for the presentation of static or dynamic phosphene images \cite{fornos2005simulation,vurro2014simulation,li2017real}. This methodology allow controlled evaluation of normally sighted subject response and task performance which is fundamental to know the way humans perceive and interpret phosphenized renderings. SPV also offers the advantage of adapting implant designs to improve the perceptual quality using image processing techniques without involving implanted subjects. In our case, the participants were normal sighted subjects seated on a chair facing a computer screen at 1m distance resulting in a 20 degrees simulated field of view, as can be seen in Fig~\ref{Fig6}.

\begin{figure}[!h]
\begin{adjustwidth}{-2.25in}{0in} 
\centering
\includegraphics[width=0.96\textwidth]{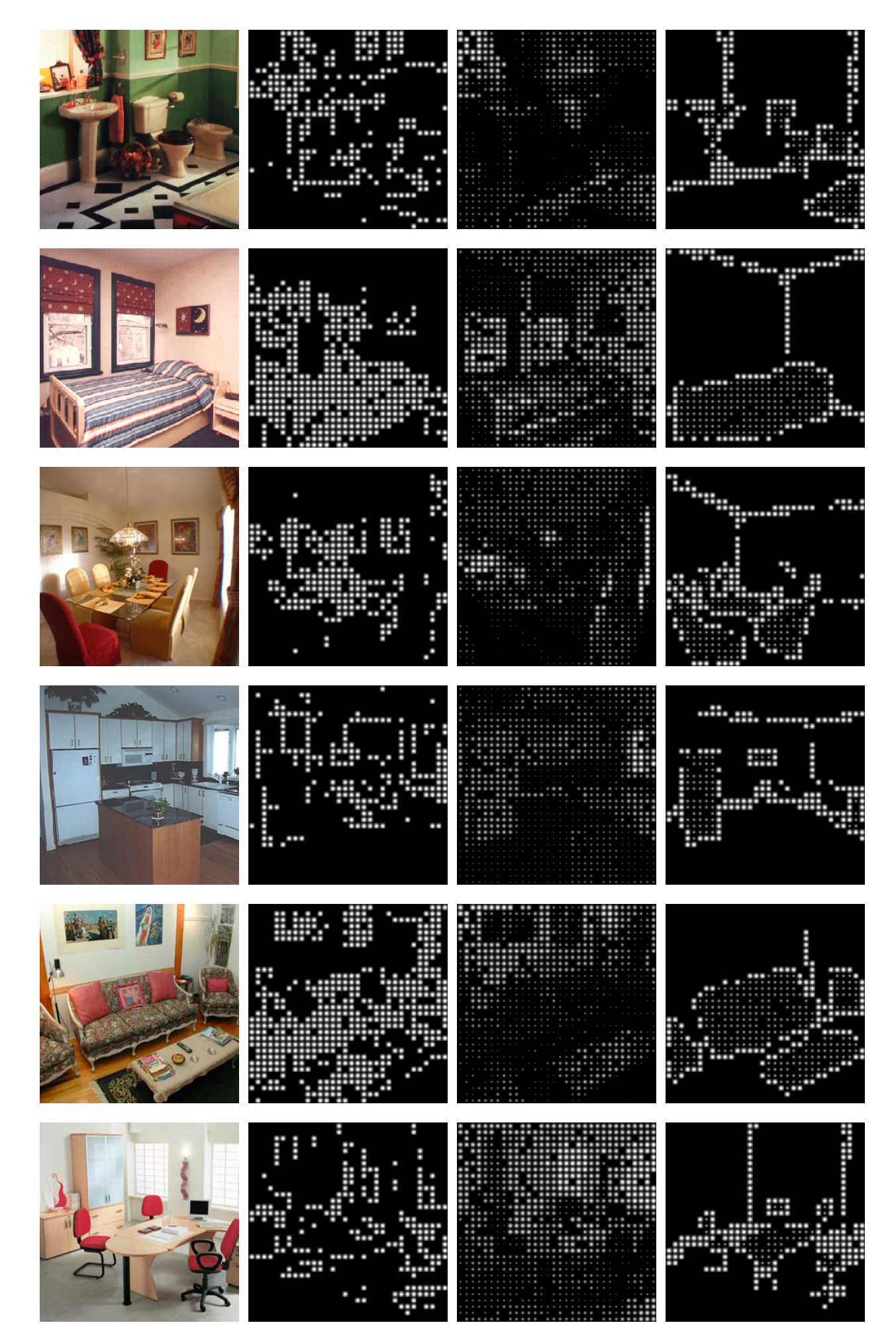}
\caption{{\bf Examples of stimuli used in the experiment.} Six examples of indoor environments represented with 1024 phosphenes (rows: bathroom, bedroom, dining room, kitchen, living room and office, respectively). Each column shows: a) input images, b) images processed using the Edge method, c) images processed using the Direct method and d) images processed by our SIE-OMS method, respectively.}
\label{Fig7}
\end{adjustwidth}
\end{figure}

For the formal experiment, subjects were recruited to complete two tasks: object recognition and room identification. The recognition accuracy was analyzed after the trials. Each trial consisted of a sequence of images presented randomly to the subject with the proposed SIE-OMS stimuli method and the two baseline methods (Edge, Direct), as can be seen in Fig~\ref{Fig7}. At the beginning of the experiment, a white dot was displayed in the center of the screen indicating where the subjects had to maintain the fixation sight until the beginning of the task. Next, each phosphenic image appeared for 10 seconds and switched for the next image automatically. This procedure was repeated for the other test images. To avoid distractions in the participants, they verbally indicate the type of objects seen in each image and their selection of room type keeping the fixation sight on the screen. The responses of each image were annotated by the experimenter. If the subjects did not respond within the 10 seconds that the image is displayed, the result of the test image was considered not answered (NA). If the subjects were only able to respond to one of the two phases of the experiment, only the unanswered phase was considered as not answered. Every 15 test images we made a pause of 30 seconds. The complete experiment took approximately 15 minutes.

\begin{figure}[!h]
\centering
\includegraphics[width=1\textwidth]{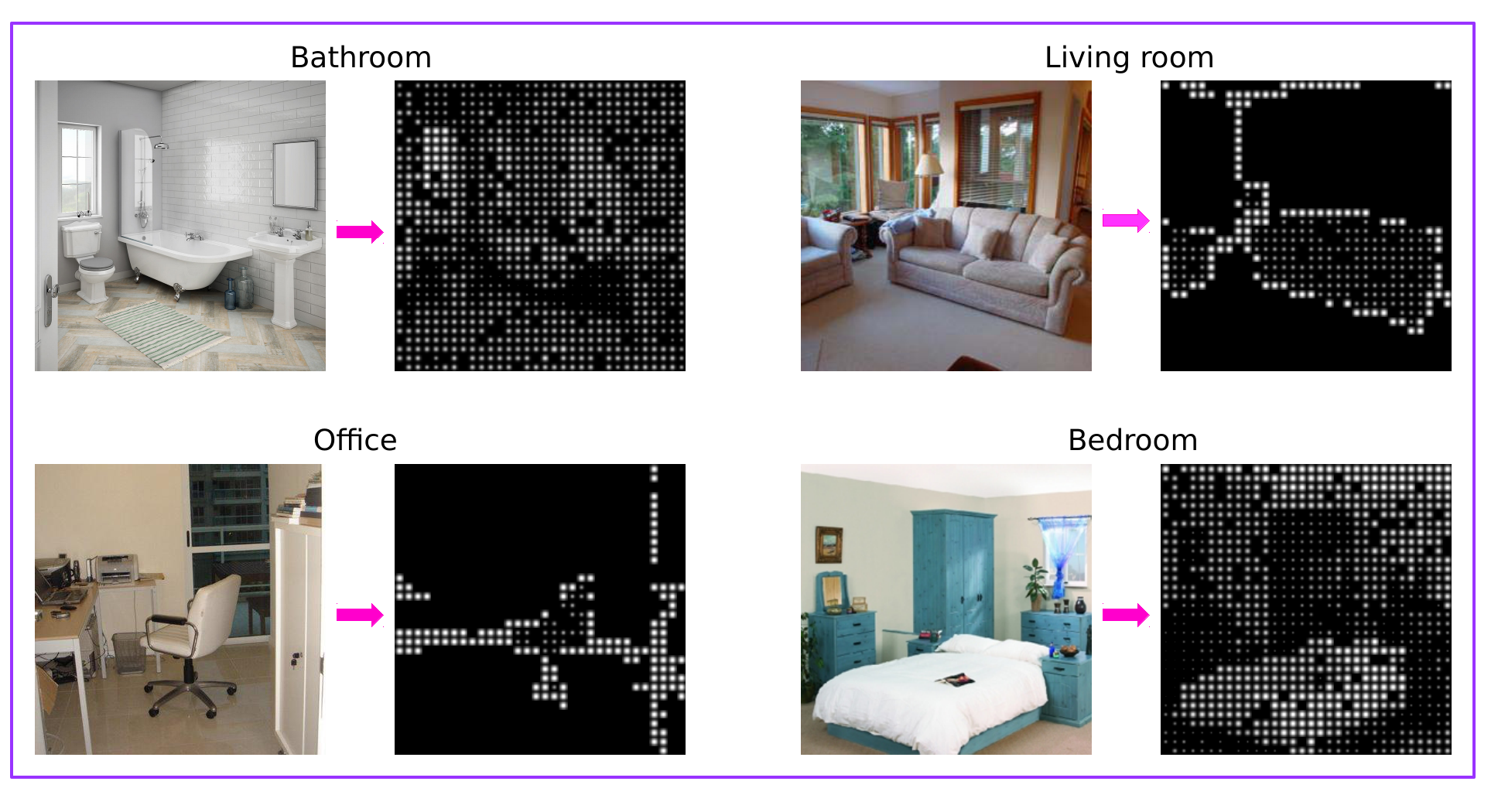}
\caption{{\bf Examples during the experiment.} Images presented at the beginning of the experiment to illustrate the kind of images they were going to see.}
\label{Fig8}
\end{figure}

The experiments were conducted using a public database of indoor scenes \cite{quattoni2009recognizing}. All the images from the database are still life scenes, from arbitrary scenarios, locations, clutter, cameras and lightning conditions. Some images are from old phone cameras with very poor quality and resolution to be more challenging as a computer vision benchmark. Thus, we replaced some images with the first results of querying Google Images with the room label, that also matched the database features (e.g., still life, mid-wide view...). For each of the six categories, we randomly selected 50 images. Hence, we conducted the experiment using 300 images from different indoor environments. The original images were processed using our proposed method and the two baseline approaches, resulting in 900 phosphenic images. Prior to beginning the experiment, subjects were informed about the number of images in the experiment (54 images per subject). Subjects were unaware that multiple image processing strategies were used in the experiment, although a screen with four images were shown to the subjects at the beginning of the test as shown in Fig. \ref{Fig8}. These demo images were not included in the experiment, to avoid learning effects.  Subjects were informed that all scenes were indoor scenarios, but they were not informed about the types of room, neither the object classes, nor the number of objects in each image. The types of room studied were: \emph{bathroom, dining room, living room, kitchen, office and bedroom}. No subject identified a type of room or scene not belonging to that list. In most of the tests, the objects identified by the subjects were: \emph{chair, table, couch, toilet, bath, sink, bed, oven/microwave, refrigerator, laptop}. This coincides with the list of classes used for our SIE-OMS which was selected without looking at the database and before conducting any trial or test. However, in two images with the direct method, a couple of subjects were able to find a \emph{window} that our system did not detect because the class was not included. Furthermore, in a couple of cases a subject wrongly identified \emph{wardrobe} and \emph{door} in images containing a \emph{fridge}.

\section*{Results}
\label{sec:results}

The following section shows the results of the experiment. We analyze separately the results of object recognition phase and room identification phase. We show the percentage of correct responses in both tasks and we include 95\% confidence intervals. We also differentiate between incorrect response and no answer.

\subsection*{Comparison of stimuli generation methods}

\begin{figure}[!h]
\includegraphics[width=1\textwidth]{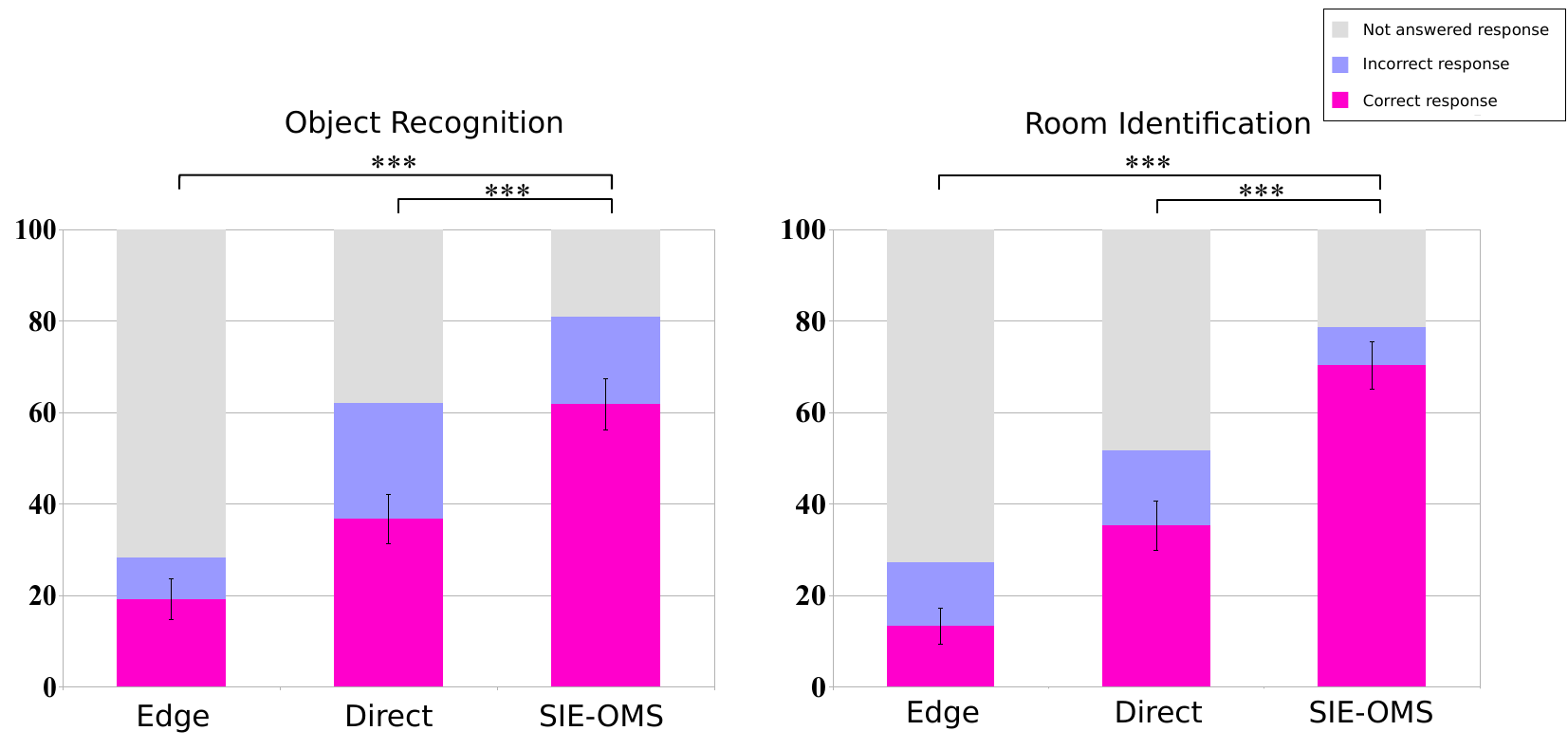}
\caption{{\bf Global results by phosphenic stimuli method.} Percentage of correct, incorrect and not answered responses in a single trial. Higher scores in correct responses indicate that subjects were able to identify and recognize the objects and the type of room in each test image. Higher ratios of not answered indicate that subjects were not able to identify and recognize the objects and the type of room in each test image. The general findings are that: SIE-OMS method improves the identification of the objects resulting to be the most effective method. This translates in an increase in the number of correct answers for the room type identification test for the SIE-OMS method. Results also show that the Edge method is the least effective with the highest percentage of non responses images for the two tasks. The test found significant difference between SIE-OM and Direct method (p$<$.001). The same conclusion was found between SIE-OM and Edge method (p$<$.001). Where: ***=p$<$.001; **=p$<$.01; *=p$<$.05; ns=p$>$.05. All t-tests paired samples, two-tailed.}
\label{Fig9}
\end{figure}

Table~\ref{table1} and Fig~\ref{Fig9} show the global results for object recognition and room identification tasks considering the proposed stimuli generator (SIE-OMS) and the two baseline methods (Edge and Direct). The analysis of the average correct responses for both tasks reveals a significant difference between methods (p$<$ 0.001). In both tasks, the results show a considerably better performance of SIE-OMS compared to the other methods. The SIE-OMS method has the highest percentage of correctly identified objects (62.78\%) compared to Edge (19.17\%) and Direct (36.83\%) methods. Likewise, there is a clear increase in the percentage of success in the room identification of SIE-OMS versus Edge and Direct method. The number of unanswered responses for our method was also smaller, indicating that there was no difficulty in the comprehension of most of the images. In contrast, it is worth noting the high percentage of unanswered responses for the Edge method, reaching more than 70\% of the scenes.

\begin{table}[!ht]
%\begin{adjustwidth}{-2.25in}{0in} % Comment out/remove adjustwidth environment if table fits in text column.
\centering
\caption{
{{\bf Global object recognition (OR) and room identification (RI) values for each phosphenic stimuli method.} Comparison of mean responses and standard deviation grouped by type of phosphenic image method (Edge, Direct and SIE-OMS). 95\% of confidence interval for the mean difference.}}
\begin{tabular}{|c|c|c|}
\hline
\textbf{Method} & \textbf{\begin{tabular}[c]{@{}c@{}}\% OR\end{tabular}} & \textbf{\begin{tabular}[c]{@{}c@{}}\% RI\end{tabular}} \\  \thickhline
Edge           & 19.17 $\pm$ 4.45                                                                    & 13.33 $\pm$ 3.85                                                                  \\ \hline
Direct    & 36.83 $\pm$ 5.46                                                                    & 35.33 $\pm$ 5.41                                                                  \\ \hline
SIE-OMS         & 62.78 $\pm$ 5.50                                                                    & 70.33 $\pm$ 5.17                                                                  \\ \hline
\end{tabular}
\begin{flushleft} 
\end{flushleft}
\label{table1}
%\end{adjustwidth}
\end{table}

\begin{figure}[!h]

\begin{adjustwidth}{-2.25in}{0in} % Comment out/remove adjustwidth environment if table fits in text column.
\centering
\includegraphics[width=1.4\textwidth]{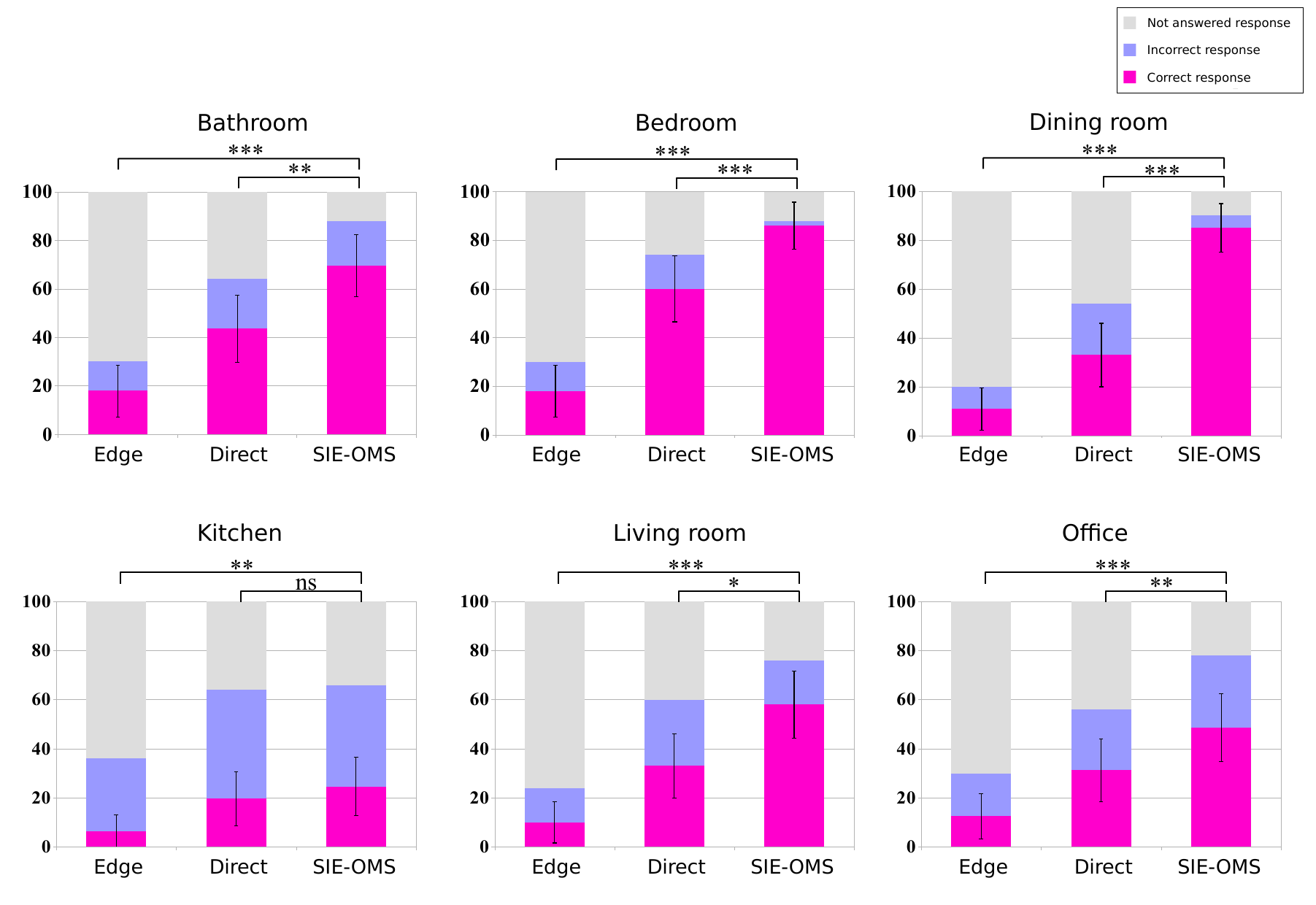}
\caption{{\bf Object recognition results for each room-type.} Higher scores in correct responses indicate that subjects were able to recognize the objects in each room. Higher ratios in non responses indicate that subjects were not able to recognize the objects in each room. The SIE-OMS method obtained the highest score of the three methods in all room types compared with Edge and Direct methods. The results also show how the most difficult room was the kitchen. ***=p$<$.001; **=p$<$.01; *=p$<$.05; ns=p$>$.05. All t-tests paired samples, two-tailed.}
\label{Fig10}
\end{adjustwidth}
\end{figure}

\begin{figure}[!h]
\begin{adjustwidth}{-2.25in}{0in} % Comment out/remove adjustwidth environment if table fits in text column.
\centering
\includegraphics[width=1.4\textwidth]{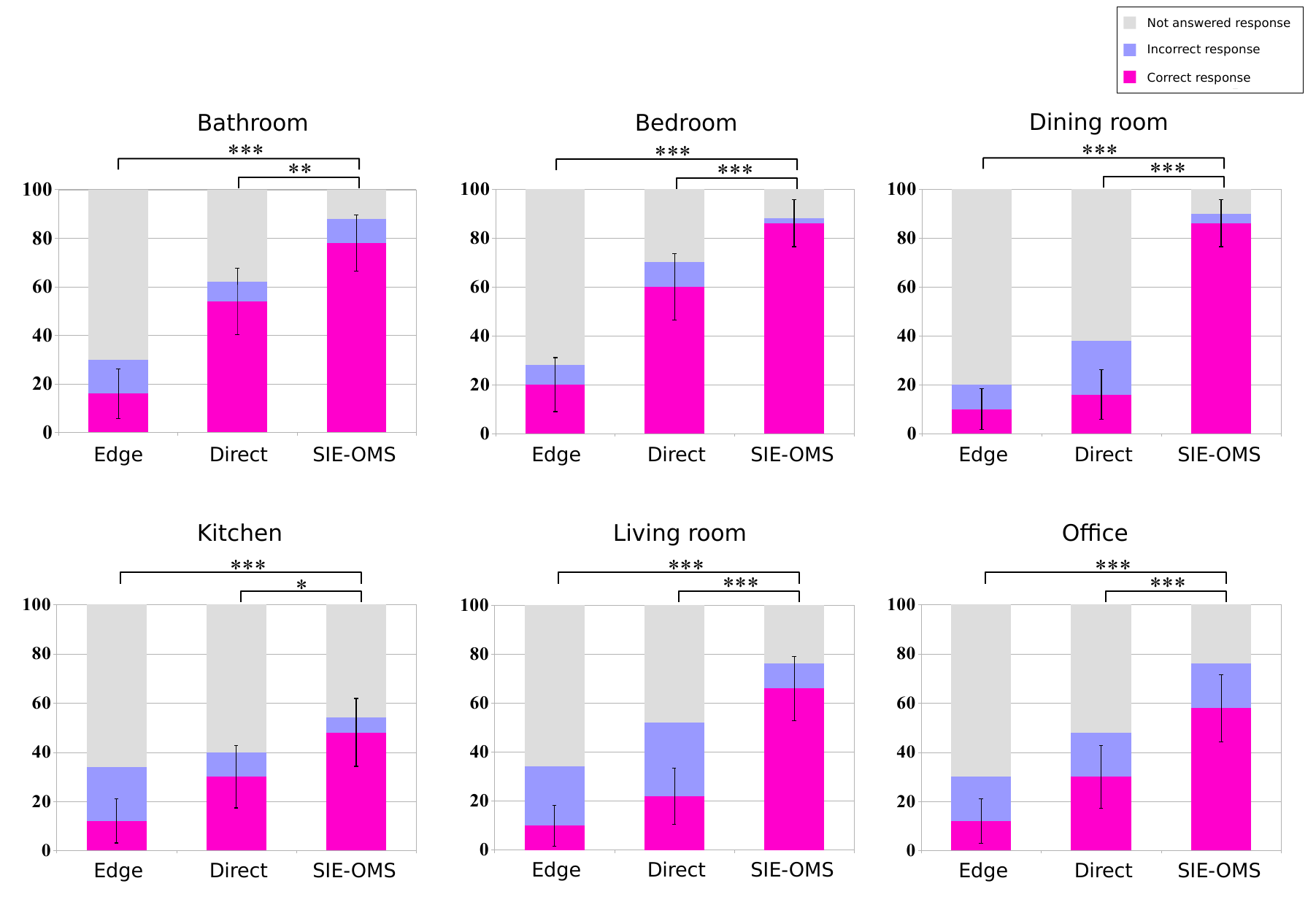}
\caption{{\bf Room identification results for each room-type.} Higher scores in correct responses indicate that subjects were able to recognize the type of room in each test image. Higher ratios in non responses indicate that subjects were not able to recognize the type of room in each image. The SIE-OMS method obtained the highest score of the three methods in all room-type compared with Edge and Direct methods. In the same way as in the identification of objects, results also showed how the most difficult room was the kitchen. ***=p$<$.001; **=p$<$.01; *=p$<$.05; ns=p$>$.05. All t-tests paired samples, two-tailed.}
\label{Fig11}
\end{adjustwidth}
\end{figure}

Fig~\ref{Fig10} and Fig~\ref{Fig11} show the results for the object recognition and room identification tasks for each room-type, respectively. As before, when comparing the baseline methods versus our approach, the highest number of correct responses is obtained for SIE-OMS method for all room types. Besides, the largest difference in results was obtained comparing the Edge method versus the SIE-OMS method (p$<0.001$) for all room-types. However, there was no significant difference for \emph{kitchen} type in Direct vs SIE-OMS (p=0.464). Similarly, there is a significant difference for \emph{living room} (p$<0.05$), \emph{office} (p$<0.01$) and \emph{bathroom} (p$<0.01$). 
On the other hand, the results of room identification task for each room-type (Fig~\ref{Fig11}) provide additional support for the SIE-OMS method since this method also has the best percentage of correct responses in each room-type, exceeding 85\% for the cases of bedroom and dining room. In the same way as in the identification of objects, the case of the \emph{kitchen} obtained the worst results, followed by the \emph{office} case.
Taken together, these findings indicate that SIE-OMS method was significantly effective improving object recognition and room identification, yet also significantly more effective than the baseline methods, Edge and Direct.

\begin{figure}[!h]
\begin{adjustwidth}{-2.25in}{0in} 
\centering
\includegraphics[width=1\textwidth]{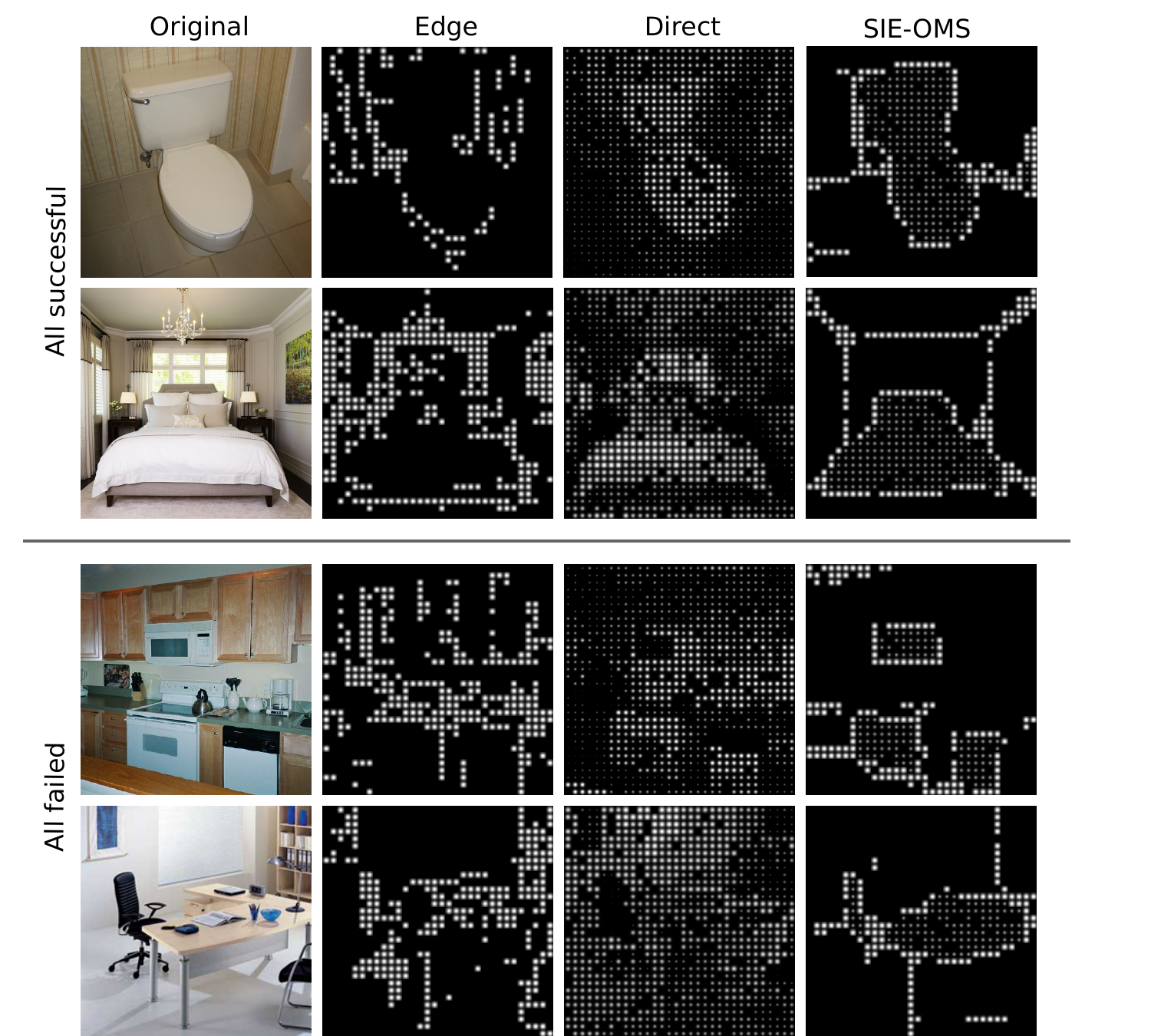}
\caption{{\bf Successful and failed images results.} Some examples of phosphenic images generated with the three methods. Successful images (top rows) and cases of images failed by the subjects (bottom rows) with the three approaches: Edge, Direct and SIE-OMS, respectively.}
\label{Fig12}
\end{adjustwidth}
\end{figure}

Fig~\ref{Fig12} shows four examples of failed and successful tests from the three methods. The two top rows show a \emph{bathroom} and a \emph{bedroom} scene where the identification of the objects and room was a success for all the methods. This is due to the location of a characteristic object with a clear silhouette in the center of the image that also helps in the identification of the room. Contrary, the bottom rows show a \emph{kitchen} and an \emph{office} where the recognition of the objects and the identification of the type of room failed in all cases as a result of the lack of distinguishable shapes (rectangle silhouettes) and visual clutter.

\subsection*{Performance analysis of SIE-OMS}

We also analyzed the performance of the proposed SIE-OMS method. The SIE-OMS system detected all the clearly visible objects of the scenes and even most of the occluded objects that matched the selected classes. Structural edges also improve the performance of our method. Recovering the main structure of the room provide sense of scale or perspective of the objects and hence a better understanding of the 3D scene. Table~\ref{table2} shows the confusion matrix of room-type based on answered images (correct and incorrect responses). Table~\ref{table3} shows the confusion matrix of the room-type based on the total images of the test (correct, incorrect and no answer).

\begin{table}[!ht]
\begin{adjustwidth}{-2.25in}{0in} % Comment out/remove adjustwidth environment if table fits in text column.
\centering
\caption{
{\bf Confusion matrix results for room identification based only on answered images (correct and incorrect responses) using SIE-OMS method.}}
\begin{tabular}{|c|c|c|c|c|c|c|cc}
\hline
\textbf{\begin{tabular}[c]{@{}c@{}}Actual/Predicted\end{tabular}} & \textbf{Bathroom} & \textbf{Bedroom} & \textbf{Dining room} & \textbf{Kitchen} & \textbf{Living room} & \textbf{Office} & \multicolumn{1}{c|}{\textbf{Total}} & \multicolumn{1}{c|}{\textbf{Recall}} \\ \thickhline
Bathroom                  & \textbf{0.89}     & 0.00             & 0.00                 & 0.00             & 0.09                 & 0.02            & \multicolumn{1}{c|}{1.00}           & \multicolumn{1}{c|}{\textbf{88.64}}  \\ \hline
Bedroom                   & 0.00              & \textbf{0.98}    & 0.00                 & 0.00             & 0.02                 & 0.00            & \multicolumn{1}{c|}{1.00}           & \multicolumn{1}{c|}{\textbf{97.73}}  \\ \hline
Dining room               & 0.00              & 0.00             & \textbf{0.96}        & 0.02             & 0.00                 & 0.02            & \multicolumn{1}{c|}{1.00}           & \multicolumn{1}{c|}{\textbf{95.56}}  \\ \hline
Kitchen                   & 0.04              & 0.00             & 0.00                 & \textbf{0.89}    & 0.04                 & 0.04            & \multicolumn{1}{c|}{1.00}           & \multicolumn{1}{c|}{\textbf{88.89}}  \\ \hline
Living room               & 0.05              & 0.03             & 0.00                 & 0.00             & \textbf{0.87}        & 0.05            & \multicolumn{1}{c|}{1.00}           & \multicolumn{1}{c|}{\textbf{86.84}}  \\ \hline
Office                    & 0.13              & 0.08             & 0.00                 & 0.03             & 0.00                 & \textbf{0.76}   & \multicolumn{1}{c|}{1.00}           & \multicolumn{1}{c|}{\textbf{76.32}}  \\ \hline
\textbf{Total}            & 1.11              & 1.08             & 0.96                 & 0.94             & 1.02                 & 0.90            & \multicolumn{1}{c|}{6.00}           &                                      \\ \cline{1-8}
\textbf{Precision}        & \textbf{80.02}    & \textbf{90.28}   & \textbf{100.00}      & \textbf{94.82}   & \textbf{85.21}       & \textbf{85.01}  &                                     &                                      \\ \cline{1-7}
\end{tabular}
\begin{flushleft} 
\end{flushleft}
\label{table2}
\end{adjustwidth}
\end{table}

\begin{table}[!ht]
\begin{adjustwidth}{-2.25in}{0in} % Comment out/remove adjustwidth environment if table fits in text column.
\centering
\caption{
{\bf Confusion matrix results for room identification based on the total images (correct, incorrect and no answer (NA)) using SIE-OMS method.}}
\begin{tabular}{|c|c|c|c|c|c|c|lcc}
\hline
\textbf{\begin{tabular}[c]{@{}c@{}}Actual/Predicted\end{tabular}} & \textbf{Bathroom} & \textbf{Bedroom} & \textbf{Dining.r} & \textbf{Kitchen} & \textbf{Living.r} & \textbf{Office} & \multicolumn{1}{l|}{\textbf{NA}} & \multicolumn{1}{c|}{\textbf{Total}} & \multicolumn{1}{c|}{\textbf{Recall}} \\ \thickhline
Bathroom                  & \textbf{0.78}     & 0.00             & 0.00                 & 0.00             & 0.08                 & 0.02            & \multicolumn{1}{l|}{0.12}          & \multicolumn{1}{c|}{1.00}           & \multicolumn{1}{c|}{\textbf{78.00}}  \\ \hline
Bedroom                   & 0.00              & \textbf{0.86}    & 0.00                 & 0.00             & 0.02                 & 0.00            & \multicolumn{1}{l|}{0.12}          & \multicolumn{1}{c|}{1.00}           & \multicolumn{1}{c|}{\textbf{86.00}}  \\ \hline
Dining room               & 0.00              & 0.00             & \textbf{0.86}        & 0.02             & 0.00                 & 0.02            & \multicolumn{1}{l|}{0.10}          & \multicolumn{1}{c|}{1.00}           & \multicolumn{1}{c|}{\textbf{86.00}}  \\ \hline
Kitchen                   & 0.02              & 0.00             & 0.00                 & \textbf{0.48}    & 0.02                 & 0.02            & \multicolumn{1}{l|}{0.46}          & \multicolumn{1}{c|}{1.00}           & \multicolumn{1}{c|}{\textbf{48.00}}  \\ \hline
Living room               & 0.04              & 0.02             & 0.00                 & 0.00             & \textbf{0.66}        & 0.04            & \multicolumn{1}{l|}{0.24}          & \multicolumn{1}{c|}{1.00}           & \multicolumn{1}{c|}{\textbf{66.00}}  \\ \hline
Office                    & 0.10              & 0.06             & 0.00                 & 0.02             & 0.00                 & \textbf{0.58}   & \multicolumn{1}{l|}{0.24}          & \multicolumn{1}{c|}{1.00}           & \multicolumn{1}{c|}{\textbf{58.00}}  \\ \hline
\textbf{Total}            & 0.94              & 0.94             & 0.86                 & 0.52             & 0.78                 & 0.68            & \multicolumn{1}{l|}{1.28}          & \multicolumn{1}{c|}{6.00}           &                                      \\ \cline{1-9}
\textbf{Precision}        & \textbf{82.98}    & \textbf{91.49}   & \textbf{100.00}      & \textbf{92.31}   & \textbf{84.62}       & \textbf{85.29}  &                                    &                                     &                                      \\ \cline{1-7}
\end{tabular}
\begin{flushleft}
\end{flushleft}
\label{table3}
\end{adjustwidth}
\end{table}

Concerning the performance of our method, the recall and the precision are high, reaching in some cases up to 97\% (Table~\ref{table2}). The diagonal elements show the number of correct classifications for each class. Hence, most of confusions are found in bathroom, living room and office. Office was confused with bathroom because of the similarity of shape between some chairs and toilets. In addition, office was confused by bedrooms since many of them usually have study desks in the bedrooms. There were other less relevant cases where the dining room was confused with an office since both are composed of chairs and tables. This confusion can be explained because the database is from Northamerican locations, while the subjects live in Spain where apartments commonly join the dining room and the kitchen.

Note that when the unanswered responses are taken into account (Table~\ref{table3}), the recall for the kitchen case decreased significantly (from 88.89\% to 48.00\%). This means that the kitchen room is more difficult to be identified. This low performance in the kitchen identification is mainly because the information provided turned out to be very limited in this case. For instance, ovens, microwaves and fridges with a rectangular shape masks were sometimes confused with windows, doors or wardrobes (which are object classes not considered by our system).

\section*{Discussion}

The visual information in interpretation of the phosphene simulation is an important issue due to the limited capabilities of retinal implants. To overcome the reduced resolution and limited dynamic range of implants, SPV researchers have tried to optimize the image presentation to deliver the effective visual information in daily activities. For example, Vergnieux et al. \cite{vergnieux2017simplification} limited the cues in a virtual scene using different renderings methods, highlighting structural cues such as the edges of different surfaces for navigation. For the same purpose, Perez et al. \cite{perez2017depth} proposed a phosphene map coding using a ground representation of the obstacle-free space and a ceiling representation based on vanishing lines. Wang et al. \cite{wang2014moving} proposed two image representation strategies using background subtraction to segment moving elements for object recognition. Similarly, Guo et al. \cite{guo2018optimization} and Li et al. \cite{li2018image}  proposed two image processing strategies based on a saliency segmentation technique. For scene recognition, McCarthy et al. \cite{mccarthy2013augmenting} presented a visual representation based on intensity augments in order to emphasise regions of structural change.

In terms of complex scene understanding, just few SPV studies have been proposed \cite{al2018extraspectral,li2018image}. It is well established that in realistic environment, which is made of complex scenes, the observer is forced to select relevant elements \cite{boucart2011object}. That is necessary to quickly understand the meaning of a scene as well as for object search. For instance, the set of objects in the environment give rise to a corresponding set of representations in the observer. Each representation describe the identity, location, and meaning of the item it refers to finally forming a literal representation of the environment. Some research on the visual perception of subjects has shown that because the fixation of the gaze changes in a short period of time when an environment is observed, the content of the scene can not be integrated into a complete and detailed representation \cite{biederman1987recognition, peterson2003perception}. This suggests that such complete and detailed representations are not needed to obtain a meaning of the scene. Just a few set of object and scene elements are enough to provide access to semantic information \cite{biederman2017semantics}. 

A well-known result in psychophysics highlight that grouping elements in a scene are fundamental for scene understanding \cite{hoffman1997salience}. First, the grouping of pixels in a region defines a contour. In many cases, shape alone permits recognition of objects. Biederman et al. \cite{biederman1987recognition,biederman1988surface} demonstrated that the silhouettes of the objects are generally very easy to identify and to recognize. The silhouette conveys only part of the visual information needed for the interpretation of an object. Concretely, the concepts such as convexities, concavities, or inflections of contours allow the observer to infer the surface geometry \cite{koenderink1984does}. However, this bottom-up perception can be computed first and to help any top-down search to converge to the right answer. This can help to understand the visual scene through the interpretation of its content. However, in order to fully understand a scene, it is not only important the identification of individual objects comprising the scene but also their relative locations and relations \cite{biederman2017semantics}. Based on this idea, the segmentation of the scene into elements with semantic meaning becomes a key point in low vision.

The state of the art in semantic segmentation include deep learning algorithms. Specifically, FCNs have proven to be successful in various recognition tasks such as semantic segmentation of images. In this work, we use two FCNs to select and highlight useful information in indoor scenes such as relevant object masks and silhouettes and structural edges which recover the main structure of the scene providing sense of scale or perspective of the objects. Even though deep learning methods are known for being resource-hungry during training, they can achieve real time performance for prediction even in mobile or embedded devices \cite{wang2018deep}. Thus, our method could be easily integrated in an implant device. 

The performance of the proposed visual stimuli, the SIE-OMS method, was investigated for object recognition and room identification tasks. Our results show that generating phosphene images by extracting specific segments of the scene such as structural informative edges and objects shapes are effective at improving object recognition and room identification. Moreover, the SIE-OMS method produces a large improvement on object recognition and room type identification compared to standard methods in SPV. However, object appearance alone is not enough for accurate object recognition in certain scenes. Since the only piece of visual data that our system uses for each object is its shape, the introduction of complementary information such as the object label could make recognition easier and avoid confusion between objects with similar shape. These factor will be considered in future studies for more realistic practices. We also note that structural edge detection is fundamental for performing tasks such as self-orientation and building a mental map of the environment. Finding such structure is crucial for personal mobility with visual prosthesis, where the bandwidth of image information that can be represented per frame is quite restricted. Overall, we can affirm that the perception and comprehension of the scene can be obtained with just a few set of elements represented in the environment.

\section*{Conclusion}

We present a new visual representation of indoor environments for prosthetic vision, which emphasizes the scene structure and object shapes. By combining the output of two FCN for structural informative edges and object masks and silhouettes, we have demonstrated how different scenes and objects can be quickly recognized even under the restricted conditions of prosthetic vision. Our results demonstrate that our method is well suited for indoor scene understanding over traditional image processing methods used in visual prostheses. The key idea of our current results is that, with only a few significant elements of the scene, it is possible to obtain a good perception of the environment, even in complex and occluded scenes. This work can be used to help visually impaired people to significantly improve their ability to adapt to the surrounding environment. 

\section*{Supporting information}

% Include only the SI item label in the paragraph heading. Use the \nameref{label} command to cite SI items in the text.
%\paragraph*{S1 Fig.}
%\label{S1_Fig}
%{\bf Bold the title sentence.} Add descriptive text after the title of the item (optional).

%\paragraph*{S2 Fig.}
%\label{S2_Fig}
%{\bf Lorem ipsum.} Maecenas convallis mauris sit amet sem ultrices gravida. Etiam eget sapien nibh. Sed ac ipsum eget enim egestas ullamcorper nec euismod ligula. Curabitur fringilla pulvinar lectus consectetur pellentesque.

\paragraph*{S1 Appendix.}
\label{S1_File}
{\bf Supplementary material.} 

%\paragraph*{S2 Dataset.}
%\label{S2_File}
%{\bf Examples of phosphenic images for the three methods SIE-OMS, Edge and Direct.} 

%\paragraph*{S1 Video.}
%\label{S1_Video}
%{\bf Lorem ipsum.}  Maecenas convallis mauris sit amet sem ultrices gravida. Etiam eget sapien nibh. Sed ac ipsum eget enim egestas ullamcorper nec euismod ligula. Curabitur fringilla pulvinar lectus consectetur pellentesque.

%\paragraph*{S1 Appendix.}
%\label{S1_Appendix}
%{\bf Lorem ipsum.} Maecenas convallis mauris sit amet sem ultrices gravida. Etiam eget sapien nibh. Sed ac ipsum eget enim egestas ullamcorper nec euismod ligula. Curabitur fringilla pulvinar lectus consectetur pellentesque.

%\paragraph*{S1 Table.}
%\label{S1_Table}
%{\bf Lorem ipsum.} Maecenas convallis mauris sit amet sem ultrices gravida. Etiam eget sapien nibh. Sed ac ipsum eget enim egestas ullamcorper nec euismod ligula. Curabitur fringilla pulvinar lectus consectetur pellentesque.

\section*{Acknowledgments}
This work was supported by project DPI2015-65962-R (MINECO/FEDER, UE) and BES-2016-078426 (MINECO).

%\nolinenumbers

% Either type in your references using
% \begin{thebibliography}{}
% Text
% \end{thebibliography}
%
% or
%
% Compile your BiBTeX database using our plos2015.bst
% style file and paste the contents of your .bbl file
% here. See http://journals.plos.org/plosone/s/latex for 
%\bibliographystyle{plos2015}
\bibliography{bibliography}

%\begin{thebibliography}{10}

%\bibitem{bib1}
%Conant GC, Wolfe KH.
%\newblock {{T}urning a hobby into a job: how duplicated genes find new
%  functions}.
%\newblock Nat Rev Genet. 2008 Dec;9(12):938--950.

%\bibitem{bib2}
%Ohno S.
%\newblock Evolution by gene duplication.
%\newblock London: George Alien \& Unwin Ltd. Berlin, Heidelberg and New York:
%  Springer-Verlag.; 1970.

%\bibitem{bib3}
%Magwire MM, Bayer F, Webster CL, Cao C, Jiggins FM.
%\newblock {{S}uccessive increases in the resistance of {D}rosophila to viral
%  infection through a transposon insertion followed by a {D}uplication}.
%\newblock PLoS Genet. 2011 Oct;7(10):e1002337.

%\end{thebibliography}

\end{document}